# Using Ants as a Genetic Crossover Operator in GLS to Solve STSP

Hassan Ismkhan
Computer Department
University of Isfahan
Isfahan, Iran
H.Ismkhan@eng.ui.ac.ir

Kamran Zamanifar
Computer Department
University of Isfahan
Isfahan, Iran
zamanifar@eng.ui.ac.ir

*Abstract—* **Ant Colony Algorithm (ACA) and Genetic Local Search (GLS) are two optimization algorithms that have been successfully applied to the Traveling Salesman Problem (TSP). In this paper we define new crossover operator then redefine ACA's ants as operate according to defined crossover operator then put forward our GLS that uses these ants to solve Symmetric TSP (STSP) instances.**

## I. INTRODUCTION

GLSs are metaheuristics that not only exploit Genetic Algorithm (GA) but also use profits of local search and have been successfully applied to TSP [8-9-10-11-12].

ACA is another metaheuristic that was firstly invented by three Italy scholars. This algorithm is inspired by the behavior of real ants that enables them to find the shortest paths between food sources and their nest. Many versions of ACA have been successfully applied to TSP [13-14-16-17-18].

Each of GLS and ACA has been successfully applied to TSP but mach of ACA with GA family can be better to solve TSP [21]. Reference [15] embeds ACA with GA to solve TSP. In this paper we will combine ACA with GLS to solve TSP. we will use ACA's ants as crossover operator in GLS. Many heuristic crossovers have been invented [1-2-3-4-5-6-7] also here we presented a heuristic crossover. In this paper we design ant that operates as our heuristic crossover then we use it in our GLS.

This paper organized as follows. Next section introduces ACA and GLS. In section III we present our ant based GLS. In section IV we put forward our heuristic crossover and explain how our designed ant operates as our heuristic crossover. We state our local searches in section V. Section VI presents results of experiments. Section VII summarizes this paper.

## II. REVIEW OF ACA AND GLS

### A. ACA

ACA is inspired by the behavior of real ants that enables them to find the shortest paths between food sources and their nest. In the process of finding the shortest path the ants are guided by exploiting pheromone that each ant deposits on the ground when walking. To solve TSP, ACA executes multiple iterations. During each iteration, m ants located on different nodes beginning to build a tour in n (=problem size) times. In each time witch of ants selects next node according to transition rule then update pheromone locally. After each ant's tours are completed global pheromone update is executed [13-14]. General pseudo code for ACA is shown in Fig. 1.

```
1) for i=1 to Number-of-Itreation
2)     for k=1 to N(=number of city)
3)         while (tour_k is incomplete)
4)             ant[k] select next city (next edge) according to transition rule
5)             update pheromone locally
6)         update pheromone globally
```

Figure 1. ACA pseudo code

Ant Colony System (ACS) [14] is one of ACA versions that transition rule, local update and global update are defined as follow: transition rule uses (1)

$$S = \begin{cases} \arg\max_{u \in J_k(r)}\{[\tau(r,u)].[\eta(r,u)]^\beta\}, & \text{if } q \leq q_0 \\ \text{use } (2) & \text{, otherwise} \end{cases} \quad (1)$$

$$p_k(r,s) = \begin{cases} \dfrac{[\tau(r,s)].[\eta(r,s)]^\beta}{\sum_{u \in J_k(r)}[\tau(r,s)].[\eta(r,s)]^\beta} & \text{if } s \in J_k(r) \\ 0 & otherwise \end{cases} \quad (2)$$

When ant k that is located on node r selects city s as next city, pheromone local update is executed as (3)
$$\tau(r,s) \leftarrow (1-\rho).\tau(r,s) + \rho.\tau_0 \quad (3)$$

In ACS global updating rule uses best tour that is produced by one of ants in steps 3 to 4 of Fig. 1. Global updating rule is as (4)
$$\tau(r,s) \leftarrow (1-\alpha).\tau(r,s) + \alpha.(best\ tour\ length)^{-1} \quad (4)$$

In (1), (2), (3) and (4) $J_k(r)$ is the set of cities that remain to be visited by ant k positioned on city r and $\eta(r,u) = 1/distance(r,u)$ and q is a random number uniformly distributed in [0 ... 1], q0 ($0 \leqslant q0 \leqslant 1$), α, β and ρ are parameters and τ is pheromone array.

Max-Min Ant System (MMAS) is another version of ACA that limits pheromone values between $\tau_{min}$ and $\tau_{max}$ [16-17-18].

### B. GLS

GLS is another metaheuristic that is obtained from combination of GA and Local Search (LS) and can be implemented as Fig. 2. Demonstrated "while" loop in line 2 repeated until no better child produced. If one of produced children is better than one of population individual then "while" loop will be continued.

```
1) initialize population with a construction heuristic
2) while population is changed
3)     for i=1 to Generation-Size
4)         Select father and mother from population
5)         child ← operate crossover on father and mother
6)         operate local search on child
7)         add child to population
8)     reduce population
9) return best individual of population
```

Figure 2. General definition of GLS

[8-9-10-11-12] use GLS to solve TSP. In these papers procedure like Fig 2 is applied on initial population. In [11] 2-opt tour improvement is applied on initial population.

### III. OUR ANT BASED GLS

We use ACA's ants as crossover operator in our GLS. As crossover operator each ant takes two parents and produces child. Fig 3 shows our ant based GLS.

```
1) initialize population with a construction heuristic
2) use Classify_based_LS to improve population individuals
3) while population is changed
4)     for k=1 to Generation-Size
5)         Select father and mother from population
6)         for i=1 to N(=number of city)
7)             child_k[i] ← ant_k go to next city
8)             update τ(child_k[i-1],child_k[i]) according to (3) locally
9)         use our 2-opt and 3-opt move based local search on child_k
10)        add child_k to population
11)    reduce population
12)    use population's best individual to update pheromone globally use (4)
13) return best individual of population
```

Figure 3. Our ant based GLS

Each ant generates child in N(=number of cites) steps. In each step each ant selects next city and updates pheromone, according to (3), locally. After N steps ants produce their children and our local searches are applied on ant's tour and these tours are added to population and "for" loop in line 5 of Fig. 3 is terminated. If one of produced children is better than one of population individuals then "while" loop will be continued.

### IV. OUR ANT

#### A. Our Crossover

Our crossover operator selects a random city "c" and copies it to child at first then puts three (or more) pointers on each of father and mother (one of pointer in each parent must be pointed to right neighbor of selected city) then probes witch of pointed nodes is nearest to "c" then it is copied to "c" and its pointer goes one forward. Please consider that when witch of pointers reach to beginning position of other pointers then it must be removed. For example suppose that distance array of graph will be as Fig. 4, Fig. 5 show two steps of our crossover operation.

This crossover uses pointers to operate so we call it pointer based crossover (PBX).

|   | 1 | 2 | 3 | 4 | 5 | 6 | 7 | 8 |
|---|---|---|---|---|---|---|---|---|
| 1 | 0 | 12 | 19 | 31 | 22 | 17 | 23 | 12 |
| 2 | 12 | 0 | 15 | 37 | 21 | 28 | 35 | 22 |
| 3 | 19 | 15 | 0 | 50 | 36 | 35 | 35 | 21 |
| 4 | 31 | 37 | 50 | 0 | 20 | 21 | 37 | 38 |
| 5 | 22 | 21 | 36 | 20 | 0 | 25 | 40 | 33 |
| 6 | 17 | 28 | 35 | 21 | 25 | 0 | 16 | 18 |
| 7 | 23 | 35 | 35 | 37 | 40 | 16 | 0 | 14 |
| 8 | 12 | 22 | 21 | 38 | 33 | 18 | 14 | 0 |

Figure 4. Distance array of graph (dis)

| Selected node (c): 4 (is selected randomly) | Selected node (c): 5 |
|---|---|
| Father:  4  5  7  3  1  2  6  8<br>         ↑  ↑     ↑<br>   F_pointer 1  F_pointer 2  F_pointer 3 | Father:  4  5  7  3  1  2  6  8<br>            ↑     ↑<br>      F_pointer 2  F_pointer 3 |
| Mother:  3  1  7  5  6  4  2  8<br>         ↑  ↑  ↑<br>   M_pointer 1  M_pointer 2  M_pointer 3 | Mother:  3  1  7  5  6  4  2  8<br>            ↑  ↑  ↑<br>      M_pointer 1  M_pointer 2  M_pointer 3 |
| child:  4<br>dis[4,5] is shorter than dis[4,7], dis[4,1] and dis[4,6] so F_pointer1 must go one forward | child:  4  5<br>F_pointer1 reaches to F_pointer2 so must be removed |
| Selected node (c): 5 (winner node of previous step) | Selected node (c): 1 |
| Father:  4  5  7  3  1  2  6  8<br>            ↑     ↑<br>      F_pointer 2  F_pointer 3 | Father:  4  5  7  3  1  2  6  8<br>               ↑           ↑<br>         F_pointer 2     F_pointer 3 |
| Mother:  3  1  7  5  6  4  2  8<br>         ↑  ↑  ↑<br>   M_pointer 1  M_pointer 2  M_pointer 3 | Mother:  3  1  7  5  6  4  2  8<br>         ↑  ↑     ↑<br>   M_pointer 1  M_pointer 2  M_pointer 3 |
| child:  4  5<br>dis[5,1] is shorter than dis[5,7] and dis[5,6] so F_pointer3 must go one forward | child:  4  5  2 |

Figure 5. Two first steps of our crossover

## B. Our Ant

Our ant is same to pointed crossover; it takes father and mother and in each step adds one city to child according to transition rule that is defined in (5)

$$S = \begin{cases} \arg\max_{u \in PC}\{[\tau(r,u)].[\eta(r,u)]^\beta\}, & \text{if } q \leq q_0 \\ \text{use (6)} & \text{, otherwise} \end{cases} \quad (5)$$

$$p_k(r,s) = \begin{cases} \dfrac{[\tau(r,s)].[\eta(r,s)]^\beta}{\sum_{u \in PC}[\tau(r,s)].[\eta(r,s)]^\beta} & \text{if } s \in PC \\ 0 & \text{otherwise} \end{cases} \quad (6)$$

PC is union of sets of pointed city in father and mother.

## V. OUR LOCAL SEARCHES

### A. 2opt_move_based_LS

2-opt move deletes two edges and reconnects two subpaths in other way [19] (see Fig. 6). In Fig. 5 can be easily seen that 2opt_move selects a subpath from end and reverses it and the cost of obtained tour can be calculated with (1):

$$\text{Cost}_{\text{Obtained tour}} = \text{Cost}_{\text{Original tour}}$$

$$-\text{Distance}(A_{End}, B_{First}) - \text{Distance}(B_{End}, A_{First})$$

$$+ \text{Distance}(A_{End}, B_{End}) + \text{Distance}(B_{End}, A_{First}) \quad (7)$$

2opt_move_based_LS uses 2-opt-move showed in Fig. 6. 2-opt-move first selects one point at random and considers the subpath from selected point to end of tour then calculates Predicted_Cost with (7), if Predicted_Cost is be small than tour's cost then subpath is inversed. 2opt_move_based_LS repeat 2-opt-move until Predicted_Cost is been greater than tour cost.

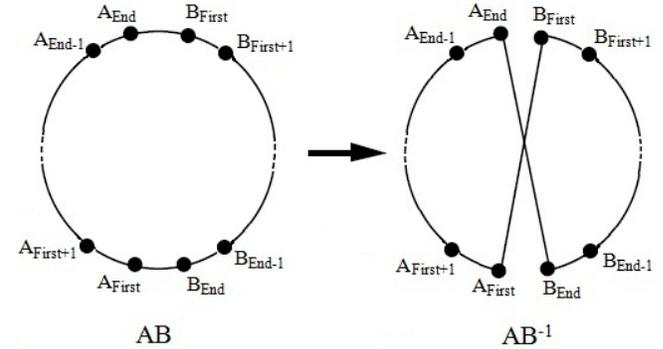

Figure 6. 2-opt move: we use X, $X^{-1}$ replace of ($X_{First}X_{First+1} \ldots X_{End-1}X_{End}$) and ($X_{End} X_{End-1} \ldots X_{First+1} X_{First}$)

### B. 3opt_move_based_LS

3-opt move deletes three edges and reconnects three subpaths in other ways [19] (see Fig. 7). The cost of obtained tours can be calculated with (8):

$$\text{Cost}_{\text{Obtained tour}} = \text{Cost}_{\text{Original tour}}$$
$$- \text{Deleted edges weight}$$
$$+ \text{Added edges weight} \quad (8)$$

For example in method 1 of Fig. 7 deleted and added edges sets are $\{(A_{End}, B_{First}), (B_{End}, C_{First}), (C_{End}, A_{First})\}$ and $\{(A_{End}, C_{First}), (C_{End}, B_{First}), (B_{End}, A_{First})\}$ respectively.

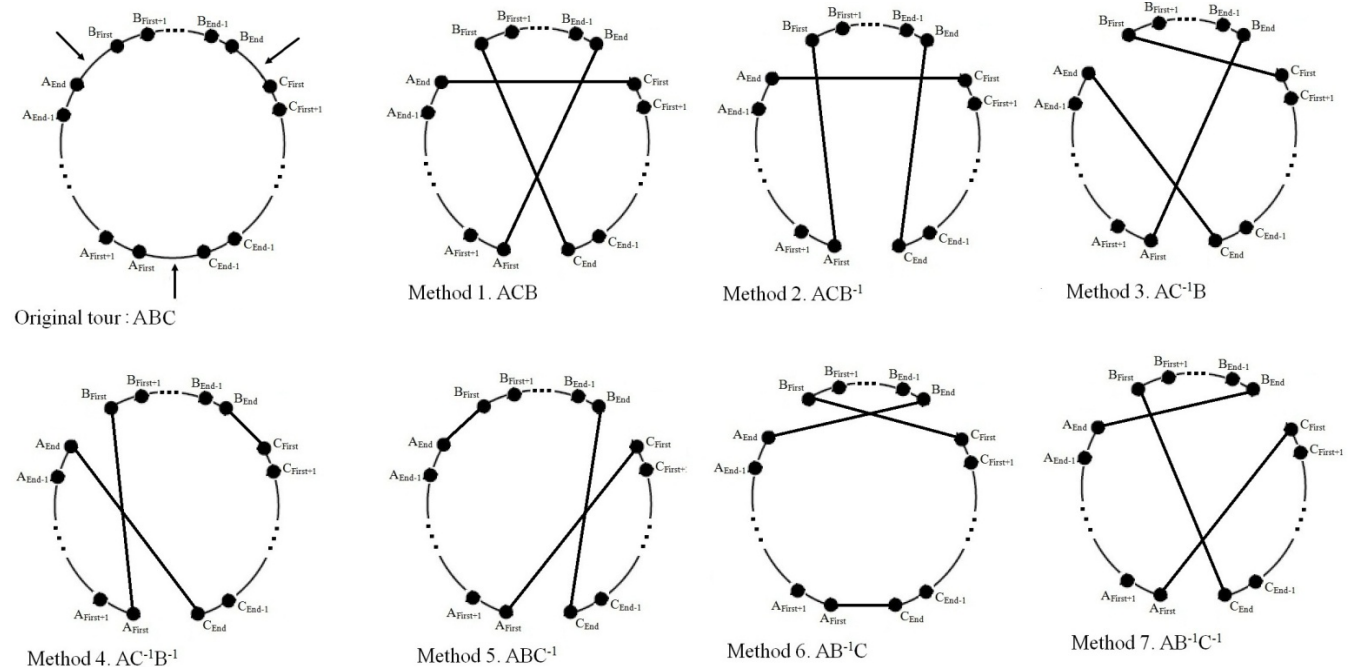

Figure 7. 3-opt move: seven distinct tour are obtained from original tour. We use X, $X^{-1}$ replace of ($X_{First}X_{First+1} \ldots X_{End-1}X_{End}$) and ($X_{End} X_{End-1} \ldots X_{First+1} X_{First}$)

3-opt move uses (8) and predicts that witch of methods in Fig. 7 decreases tour cost more than other then operates according to it. If neither of methods produce a better tour then 3-opt move returns an input tour as an output. 3opt_move_based_LS repeats 3-opt move until no better tour is produced.

## C. Classify_based_LS

This local search uses classify method that operates as follow: chooses first node of tour and copies it to variable "c" and also puts two pointer on tour, first pointer pointes to second position of tour and second pointer pointes to random position of tour then probes witch of pointed nodes is nearest to "c" then its pointer go forward and it comes to right neighbor of "c" and also it is copied to "c" and this process repeats until all nodes are probed. Repetition of this method lead to nodes that are near to first node of tour are collected in left hand side of tour also other that are far from first node are collected in right hand side of tour. Our Classify_based_LS repeats this method for K (fixed number) time or until output tour cost does not decrease. Classify method has drawback when nearest nodes to first node are located in left side of tour, example in Fig. 8 shows this problem. This example uses distance array of Fig. 4. In next section we show that this method is partially effective when it works with tours that are constructed randomly and it can decrease cost of random tour up to 65% so we use Classify_based_LS to improve initial population individuals.

| Step 0: | Step 1: c=4, dis[4,5] <dis[4,7] |
| input 4 5 1 2 7 6 3 8 | input 4 5 1 2 7 6 3 8 |
| | ↑ ↑ |
| | Pointer 1 pointer 2 |
| output 4 | output 4 5 |
| Step 2: c=2, dis[5,1] < dis[5,7] | Step 3: c=1, dis[1,2] > dis[1,7] |
| input 4 5 1 2 7 6 3 8 | input 4 5 1 2 7 6 3 8 |
| ↑ ↑ | ↑ ↑ |
| Pointer 1 pointer 2 | Pointer 1 pointer 2 |
| output 4 5 1 | output 4 5 1 2 |
| In step 3 pointer 1 reaches to pointer 2 and must be removed so pointer 2 remains alone then remaining nodes are copied to output respectively and output is same to input. ||

Figure 8. Classify method drawback

## VI. IMPLEMENTION AND RESULTS

We implemented all of algorithms with language c# and used .NET 2008 also we ran all experiments on AMD Dual Core 2.6 GHZ. In this section we show effect of Classify_based_LS on improving of tours that are constructed randomly also we put forward results for our ant based GLS. We use STSP benchmarks in our experiments. These benchmarks are eil51, eil76, kroA100 and a280 [20].

### A. effect of Classify_based_LS on improving of random tours

In this experiment we applied Classify_based_LS on 30 random tours of each mentioned instances for 30 times. The results of this experiment are given in Table I. First and second rows are calculating according to (9) and (10) respectively and third row, "Time" shows average running time of Classify_based_LS when applying to each of tours.

$$Percent\ of\ improving\ in\ average = \left[\frac{Average\ cost\ of\ initial\ tours\ -\ Average\ cost\ of\ tours\ after\ applying\ Classify\_based\_LS}{Average\ cost\ of\ initial\ tours}\right] \times 100 \quad (9)$$

$$Best\ percent\ of\ improving\ = \left[\frac{Cost\ of\ tour\ befor\ appliying\ Classify\_based\_LS\ -\ Cost\ of\ tour\ after\ appliying\ Classify\_based\_LS}{Cost\ of\ tour\ befor\ appliying\ Classify\_based\_LS}\right] \times 100 \quad (10)$$

TABLE I. RESULTS FOR CLASSIFY_BASED_LS

| | eil51 | Eil76 | kroA100 | A280 |
|---|---|---|---|---|
| Percent of improving in average | 40.93 | 44.95 | 55.64 | 59.11 |
| Percent of improving at best case | 56.76 | 65.6 | 62.48 | 65.23 |
| Time | 0.001 | 0.002 | 0.003 | 0.008 |

TABLE I shows Classify_based_LS is partially effective when it works with tours that are constructed randomly and it can decrease cost of random tour up to 65%. Please consider that in this experiment Classify_based_LS repeated classify method for 4 times.

### B. Ant based GLS results

We tested our ant based GLS thirty times on each of mentioned instances. In this experiment we initialized population with 50 random individuals (population size=50) and set Generation-Size =500 also we set ACA parameters as

α=.9, β=2, ρ=.1 q0=.9. TABLE II shows results of this experiment.

TABLE II.  EXPRIMENTAL RESULTS OF ANT BASED GLS FOR SOME TSPLIB INSTANCES, GENERATION SIZE=500, POPULATION SIZE=50

|  | Best length (quality) | Average length (quality) | Worst length (quality) | Average Time (second) |
|---|---|---|---|---|
| eil51 | 427 (0.23%) | 428.7 (0.63%) | 433 (1.64%) | 10.25 |
| eil76 | 543 (0.93%) | 548.07 (1.87%) | 558 (3.72%) | 16.13 |
| kroA100 | 21331 (0.23%) | 21903.67 (2.92%) | 23106 (8.57%) | 28.52 |
| A280 | 2609 (1.16%) | 2703.23 (4.82%) | 2832 (9.81%) | 133.99 |

In this table "Best length", "Average length" and "Worst length" show the best, average, and worst tour lengths of runs, respectively. "Average Time" column gives the average running time in seconds. In "Best length", "Average length" and "Worst length" columns the values in parentheses is result of calculating

$$\frac{\text{cost of solution found} - \text{known optimum cost}}{\text{known optimum cost}} \times 100$$

## VII. CONCLUSION

In this paper we presented GLS that was using ants as crossover operator. These ants were operating as our presented heuristic crossover. Our heuristic crossover was using several pointers to operate so we called it pointer based crossover (PBX). We also presented Classify_based local search to improve initial population individual. Experimental results have shown that Classify_based local search can decrease cost of random tours up to 65%.